\documentclass[conference]{IEEEtran}
\usepackage{cite}
\usepackage{amsmath,amssymb,amsfonts}
\usepackage{algorithmic}
\usepackage{graphicx}
\usepackage{textcomp}
\usepackage{xcolor}
\def\BibTeX{{\rm B\kern-.05em{\sc i\kern-.025em b}\kern-.08em
    T\kern-.1667em\lower.7ex\hbox{E}\kern-.125emX}}

\begin{document}

\title{Group Recommendation Techniques for \\Feature Modeling and Configuration
}
\author{Viet-Man Le\\
\IEEEauthorblockA{\textit{Graz University of Technology, Austria} \\
vietman.le@ist.tugraz.at}}

\maketitle

\begin{abstract}
In large-scale feature models, feature modeling and configuration processes are highly expected to be done by a group of stakeholders. In this context, recommendation techniques can increase the efficiency of feature-model design and find optimal configurations for groups of stakeholders. Existing studies show plenty of issues concerning feature model navigation support, group members' satisfaction, and conflict resolution. This study proposes group recommendation techniques for feature modeling and configuration on the basis of addressing the mentioned issues.
\end{abstract}


\section{research problem}

Feature modeling and configuration are two development processes of Software Product Line Engineering (\textsc{SPLE}) paradigm \cite{Apel2013,pohl2005software}. Feature modeling processes specify feature models that describe the commonality and variability properties of software artifacts \cite{pohl2005software}. In feature model configuration, software applications in product line are built by selecting a set of features based on stakeholders' requirements and constraints implied in the feature model \cite{Bagheri2010}. When the product line model is large, performing the mentioned processes becomes difficult, error-prone, and time-consuming \cite{edded2019}. Moreover, it can be very hard for a stakeholder to manage a variant-rich product line and take over a large number of configuration decisions \cite{edded2019, Mendonca2008, Kuiter2019}. In this context, a new approach, so-called \textit{collaborative modeling and configuration} \cite{edded2019,Mendonca2008,stein2014preference, Kuiter2019}, has emerged to support groups of stakeholders to jointly complete feature model design and the tasks of configuration and maintenance as well as cope with the complexity of these processes.

Recommender systems are regarded as effective tools to assist users in finding relevant items and making decisions \cite{Ricci2011}. This paper proposes recommendation techniques that support feature modeling and configuration processes performed by groups of stakeholders. These processes can benefit from recommendation techniques to determine, for instance, features to be included in a configuration, an optimal configuration for a group, or next features/constraints to be considered when navigating through the list of features/constraints.

There exist a few studies in the literature that support the mentioned processes \cite{Mendonca2008,stein2014preference,edded2019,Kuiter2019}. For instance, the authors in \cite{Czarnecki2004,Mendonca2008} present an approach that provides a pre-designed process to coordinate and assign a group of experts to configuration tasks. Stein et al. \cite{stein2014preference} propose a solution to support the multi-stakeholder configuration process, which considers individual stakeholders' preferences expressed via hard and soft constraints. The mentioned studies bring various solutions for feature modeling and configuration processes in group scenarios. However, they do not exploit the potentials coming along with the application of recommender systems. In this context, there exist the following open research issues: 


\textbf{Gap 1 - Feature model navigation support}: When working with large-scale feature models, it is tricky to identify the next features/constraints that need to be considered. Consequently, the feature model development and maintenance processes of stakeholders need the support in the navigation through the feature and constraint space. Let us assume a scenario where a group of stakeholders have already considered a small set of features/constraints. One question arising now is \textit{``how to specify the next feature/constraint to be considered from a large set of the remaining features/constraints?''}. Exploiting recommendation techniques might be a potential solution. To the best of our knowledge, there are no studies proposing recommendation approaches to support stakeholders in such a scenario.

\textbf{Gap 2 - Group members' satisfaction}: 
Stein et al. \cite{stein2014preference} propose an approach to recommend configurations to groups using \textit{social choice-based aggregation strategies} \cite{Felfernig2018a}. However, these strategies do not always generate a solution that takes into account the preferences of all group members \cite{Tran2019_UMAP_explanation}. Consequently, this triggers group members' dissatisfaction. In this context, an approach to consider \textit{``fairness aspects''} among group members can help to resolve such an issue.
    
\textbf{Gap 3 - Conﬂict resolution}: In collaborative configuration, conflicts occur when group members have different preferences for a specific feature. For instance, a group member $u_1$ wants to include feature $f_1$, whereas another group member $u_2$ doesn't. Other conflicts can be triggered when group members' preferences are inconsistent with the feature model's constraints. For instance, the inclusion of features $f_1$ and $f_2$ triggers an inconsistency since there exists a constraint $\neg(f_1 \land f_2)$ indicating that \textit{``if $f_1$ is included, then $f_2$ has to be excluded''}. Finally, conflicts can arise in the re-configuration process when stakeholders change their preferences for features or add new features. This leads to inconsistencies when aggregating the preferences of group members.
    

\section{research questions and proposed approaches}
Our goal is to propose group recommendation techniques to support collaborative feature modeling and configuration processes. To achieve this goal, we state research questions concerning the mentioned gaps and propose techniques to address them. These techniques support \textit{large-scale feature modeling and configuration scenarios}. When a feature model is huge, instead of asking group members to specify their preferences for  \textit{``all''} features, we exploit the user interaction data (collected in previous configuration processes) to predict group members' preferences for features. Before discussing our approaches, we assume the availability of user interaction data, which is used to support group members proactively. In case the data is unavailable (i.e., cold-start problems), recommendation heuristics \cite{Mazo2014,Seda2019} can be applied.

\textit{$RQ_{1}$: \textbf{How can group recommendation techniques support the selection of the next choice point? (gap 1)}}

We assume a scenario in which each group member has already visited some constraints. The order of the remaining constraints can be predicted based on, e.g.,  \textit{collaborative filtering} recommendation\cite{Ekstrand2011}. We are now interested in \textit{determining a constraint to be presented next to the group}. To address this, we aggregate the order of each constraint using an aggregation strategy (e.g., Average). The aggregated value reflects the predicted order of a constraint that the whole group should visit next. However, it could be the case that many constraints have the same predicted orders. In this context, a \textit{tie breaking rule} is needed to find the ``winner''. One possible rule is to consider the constraints' \textit{importance}. For instance, a constraint affecting many features is an important constraint and will be recommended to the group \cite{Mazo2014}.




$RQ_2$: \textit{\textbf{How to support interactive configuration processes?}}

We assume a configuration scenario where group members have articulated their preferences for a subset of features. Their preferences for the remaining features are predicted by analyzing user interaction data. From now on, group members' preferences for features are ready for the configuration process. $RQ_2$ is associated with the following sub-questions:


\textit{$RQ_{2.1}$: \textbf{How to aggregate group members' preferences in such a way that fosters fairness within the group? (gap 2)}}

To foster fairness aspects, the aggregation strategy should not ignore any group members' preferences. Our approach merges the preferences of group members when no preference conflicts occur. For instance, if the preferences of all group members for a feature is 1 (i.e., \textit{``include''} the feature to the configuration), then the group preferences for this feature is also 1. If any preference conflicts arise, a discussion phase is then triggered (see $RQ_{2.2}$). The discussion is done based on the \textit{Theory W's fundamental principle} \cite{Boehm1989,Boehm2001}, which is widely used in requirement prioritization and negotiation. The basic idea of this principle is to ensure win-win situations in which mutually satisfactory (win–win) sets of shared commitments are generated.



\textit{$RQ_{2.2}$: \textbf{How to solve conflicts between group members' preferences? (gap 3)}}

In situations where group members' preferences are contradictory, a consensus-making process is triggered to support group members to achieve an agreement. During the discussion, negotiation patterns can be provided to speed-up the discussion. By complying with the Theory W's principle (see also $RQ_{2.1}$), negotiation patterns help stakeholders to expand the option space and thereby create win-win situations. One example negotiation pattern can be: \textit{``We shouldn't include the feature $f_1$ since its price is higher than the budget, and the feature $f_2$ could be an alternative''.}


\textit{$RQ_{2.3}$: \textbf{How to resolve inconsistencies that occurred after aggregating group members' preferences? (gap 3)}}

After aggregating group members' preferences, inconsistencies between the aggregated preferences of a specific feature and the feature model's constraints can occur. To restore consistency, \textit{model-based diagnosis} \cite{Reiter1987} can be applied to suggest adaptations to be done by group members. This approach finds \textit{minimal diagnoses} \cite{felfernig2012} indicating how the group's preferences should be adapted. If many minimal diagnoses have been identified, the following question has to be answered: \textit{``Which of the alternative diagnoses should be recommended first to the group?''}. To answer this, for each diagnosis, we calculate the number of each group member's adaptations and then take the highest number as the total adaptation number of the whole group. The diagnosis with the lowest adaptation number is recommended. The general idea is \textit{``the lower the number of adaptations, the better the diagnosis''}.

$RQ_3$: \textit{\textbf{How to support reconfiguration processes? (gap 3)}}

Configuration processes entail situations where group members change their preferences for features or want to include additional features in the feature model. These changes can lead to inconsistencies between group members' preferences and the feature model's constraints. In this context, suitable adaptations are needed to restore consistency. The adaptation determination can be done using the approaches discussed in $RQ_{2.2}$ and $RQ_{2.3}$.

\section{evaluation plan}

We will use \textit{online} and \textit{offline} methods to evaluate the proposed solutions \cite{Felfernig2018Ch3}. The online methods are applied to evaluate recommended configurations' quality in terms of \textit{fairness} and \textit{user satisfaction}. Some user studies will be conducted to evaluate the effectiveness of the proposed negotiation patterns. The offline methods are utilized to evaluate the effectiveness and preciseness of recommended configurations as well as the efficiency of model-based diagnosis algorithms. Our approaches will be validated against available real-world SPLs that have been currently discussed in the literature \cite{Pereira2018, rodas-silva2019, stein2014preference, edded2020}. 
Finally, we will develop a prototype using the proposed approaches by extending a state-of-the-art tool \textsc{FeatureIDE}\cite{Thum2014}, where we will conduct in-depth evaluations of the prototype's performance.

\section{expected contributions}
Different from existing studies, we propose an approach supporting feature modeling and configuration processes in large-scale feature models. Our approach leverages group recommendation techniques and psychological models to improve the quality of chosen configurations, foster fairness aspects, and increase the satisfaction of group members with recommended configurations.


\bibliographystyle{IEEEtran}
\bibliography{main}

\begin{thebibliography}{10}
\providecommand{\url}[1]{#1}
\csname url@samestyle\endcsname
\providecommand{\newblock}{\relax}
\providecommand{\bibinfo}[2]{#2}
\providecommand{\BIBentrySTDinterwordspacing}{\spaceskip=0pt\relax}
\providecommand{\BIBentryALTinterwordstretchfactor}{4}
\providecommand{\BIBentryALTinterwordspacing}{\spaceskip=\fontdimen2\font plus
\BIBentryALTinterwordstretchfactor\fontdimen3\font minus
  \fontdimen4\font\relax}
\providecommand{\BIBforeignlanguage}[2]{{%
\expandafter\ifx\csname l@#1\endcsname\relax
\typeout{** WARNING: IEEEtran.bst: No hyphenation pattern has been}%
\typeout{** loaded for the language `#1'. Using the pattern for}%
\typeout{** the default language instead.}%
\else
\language=\csname l@#1\endcsname
\fi
#2}}
\providecommand{\BIBdecl}{\relax}
\BIBdecl

\bibitem{Apel2013}
\BIBentryALTinterwordspacing
S.~Apel, D.~Batory, C.~K{\"a}stner, and G.~Saake, \emph{Feature-Oriented
  Software Product Lines: Concepts and Implementation}.\hskip 1em plus 0.5em
  minus 0.4em\relax Berlin, Heidelberg: Springer Berlin Heidelberg, 2013.
  [Online]. Available: \url{https://doi.org/10.1007/978-3-642-37521-7}
\BIBentrySTDinterwordspacing

\bibitem{pohl2005software}
K.~Pohl, G.~B{\"o}ckle, and F.~J. van Der~Linden, \emph{Software product line
  engineering: foundations, principles and techniques}.\hskip 1em plus 0.5em
  minus 0.4em\relax Springer Science \& Business Media, 2005.

\bibitem{Bagheri2010}
E.~Bagheri, T.~Di~Noia, A.~Ragone, and D.~Gasevic, ``Configuring software
  product line feature models based on stakeholders' soft and hard
  requirements,'' in \emph{Software Product Lines: Going Beyond}, J.~Bosch and
  J.~Lee, Eds.\hskip 1em plus 0.5em minus 0.4em\relax Berlin, Heidelberg:
  Springer Berlin Heidelberg, 2010, pp. 16--31.

\bibitem{edded2019}
\BIBentryALTinterwordspacing
S.~Edded, S.~B. Sassi, R.~Mazo, C.~Salinesi, and H.~B. Ghezala, ``Collaborative
  configuration approaches in software product lines engineering: A systematic
  mapping study,'' \emph{Journal of Systems and Software}, vol. 158, p. 110422,
  2019. [Online]. Available:
  \url{http://www.sciencedirect.com/science/article/pii/S0164121219301967}
\BIBentrySTDinterwordspacing

\bibitem{Mendonca2008}
M.~Mendon\c{c}a, T.~T. Bartolomei, and D.~Cowan, ``Decision-making coordination
  in collaborative product configuration,'' in \emph{Proceedings of the 2008
  ACM Symposium on Applied Computing}, ser. SAC '08.\hskip 1em plus 0.5em minus
  0.4em\relax New York, NY, USA: Association for Computing Machinery, 2008, p.
  108–113.

\bibitem{Kuiter2019}
\BIBentryALTinterwordspacing
E.~Kuiter, S.~Krieter, J.~Kr\"{u}ger, T.~Leich, and G.~Saake, ``Foundations of
  collaborative, real-time feature modeling,'' in \emph{Proceedings of the 23rd
  International Systems and Software Product Line Conference - Volume A}, ser.
  SPLC '19.\hskip 1em plus 0.5em minus 0.4em\relax New York, NY, USA:
  Association for Computing Machinery, 2019, p. 257–264. [Online]. Available:
  \url{https://doi.org/10.1145/3336294.3336308}
\BIBentrySTDinterwordspacing

\bibitem{stein2014preference}
J.~Stein, I.~Nunes, and E.~Cirilo, ``Preference-based feature model
  configuration with multiple stakeholders,'' in \emph{Proceedings of the 18th
  International Software Product Line Conference-Volume 1}, 2014, pp. 132--141.

\bibitem{Ricci2011}
\BIBentryALTinterwordspacing
F.~Ricci, L.~Rokach, and B.~Shapira, \emph{Introduction to Recommender Systems
  Handbook}.\hskip 1em plus 0.5em minus 0.4em\relax Boston, MA: Springer US,
  2011, pp. 1--35. [Online]. Available:
  \url{https://doi.org/10.1007/978-0-387-85820-3}
\BIBentrySTDinterwordspacing

\bibitem{Czarnecki2004}
K.~Czarnecki, S.~Helsen, and U.~Eisenecker, ``Staged configuration using
  feature models,'' in \emph{Software Product Lines}, R.~L. Nord, Ed.\hskip 1em
  plus 0.5em minus 0.4em\relax Berlin, Heidelberg: Springer Berlin Heidelberg,
  2004, pp. 266--283.

\bibitem{Felfernig2018a}
A.~Felfernig, M.~Atas, D.~Helic, T.~N.~T. Tran, M.~Stettinger, and R.~Samer,
  \emph{Group Recommender Systems: An Introduction}.\hskip 1em plus 0.5em minus
  0.4em\relax Springer, 2018, ch. Algorithms for Group Recommendation, pp.
  27--58.

\bibitem{Tran2019_UMAP_explanation}
T.~N.~T. Tran, M.~Atas, A.~Felfernig, V.~M. Le, R.~Samer, and M.~Stettinger,
  ``Towards social choice-based explanations in group recommender systems,'' in
  \emph{Proceedings of the 27th ACM Conference on User Modeling, Adaptation and
  Personalization}, ser. UMAP '19.\hskip 1em plus 0.5em minus 0.4em\relax New
  York, NY, USA: ACM, 2019, pp. 13--21.

\bibitem{Mazo2014}
\BIBentryALTinterwordspacing
R.~Mazo, C.~Dumitrescu, C.~Salinesi, and D.~Diaz, \emph{Recommendation
  Heuristics for Improving Product Line Configuration Processes}.\hskip 1em
  plus 0.5em minus 0.4em\relax Berlin, Heidelberg: Springer Berlin Heidelberg,
  2014, pp. 511--537. [Online]. Available:
  \url{https://doi.org/10.1007/978-3-642-45135-5\_19}
\BIBentrySTDinterwordspacing

\bibitem{Seda2019}
\BIBentryALTinterwordspacing
S.~P. Erdeniz, A.~Felfernig, R.~Samer, and M.~Atas, ``Matrix factorization
  based heuristics for constraint-based recommenders,'' in \emph{Proceedings of
  the 34th ACM/SIGAPP Symposium on Applied Computing}, ser. SAC '19.\hskip 1em
  plus 0.5em minus 0.4em\relax New York, NY, USA: Association for Computing
  Machinery, 2019, p. 1655–1662. [Online]. Available:
  \url{https://doi.org/10.1145/3297280.3297441}
\BIBentrySTDinterwordspacing

\bibitem{Ekstrand2011}
\BIBentryALTinterwordspacing
M.~D. Ekstrand, J.~T. Riedl, and J.~A. Konstan, ``Collaborative filtering
  recommender systems,'' \emph{Found. Trends Hum.-Comput. Interact.}, vol.~4,
  no.~2, p. 81–173, Feb. 2011. [Online]. Available:
  \url{https://doi.org/10.1561/1100000009}
\BIBentrySTDinterwordspacing

\bibitem{Boehm1989}
B.~W. {Boehm} and R.~{Ross}, ``Theory-w software project management principles
  and examples,'' \emph{IEEE Transactions on Software Engineering}, vol.~15,
  no.~7, pp. 902--916, 1989.

\bibitem{Boehm2001}
B.~{Boehm}, P.~{Grunbacher}, and R.~O. {Briggs}, ``Developing groupware for
  requirements negotiation: lessons learned,'' \emph{IEEE Software}, vol.~18,
  no.~3, pp. 46--55, 2001.

\bibitem{Reiter1987}
\BIBentryALTinterwordspacing
R.~Reiter, ``A theory of diagnosis from first principles,'' \emph{Artificial
  Intelligence}, vol.~32, no.~1, pp. 57 -- 95, 1987. [Online]. Available:
  \url{http://www.sciencedirect.com/science/article/pii/0004370287900622}
\BIBentrySTDinterwordspacing

\bibitem{felfernig2012}
A.~Felfernig, M.~Schubert, and C.~Zehentner, ``\BIBforeignlanguage{English}{An
  efficient diagnosis algorithm for inconsistent constraint sets},''
  \emph{\BIBforeignlanguage{English}{Artificial Intelligence for Engineering
  Design, Analysis and Manufacturing}}, vol.~26, no.~1, pp. 53--62, 2012.

\bibitem{Felfernig2018Ch3}
\BIBentryALTinterwordspacing
A.~Felfernig, L.~Boratto, M.~Stettinger, and M.~Tkal{\v{c}}i{\v{c}},
  \emph{Evaluating Group Recommender Systems}.\hskip 1em plus 0.5em minus
  0.4em\relax Cham: Springer International Publishing, 2018, pp. 59--71.
  [Online]. Available: \url{https://doi.org/10.1007/978-3-319-75067-5\_3}
\BIBentrySTDinterwordspacing

\bibitem{Pereira2018}
\BIBentryALTinterwordspacing
J.~A. Pereira, P.~Matuszyk, S.~Krieter, M.~Spiliopoulou, and G.~Saake,
  ``Personalized recommender systems for product-line configuration
  processes,'' \emph{Computer Languages, Systems \& Structures}, vol.~54, pp.
  451 -- 471, 2018. [Online]. Available:
  \url{http://www.sciencedirect.com/science/article/pii/S147784241730043X}
\BIBentrySTDinterwordspacing

\bibitem{rodas-silva2019}
J.~{Rodas-Silva}, J.~A. {Galindo}, J.~{García-Gutiérrez}, and D.~{Benavides},
  ``Selection of software product line implementation components using
  recommender systems: An application to wordpress,'' \emph{IEEE Access},
  vol.~7, pp. 69\,226--69\,245, 2019.

\bibitem{edded2020}
\BIBentryALTinterwordspacing
S.~Edded, S.~B. Sassi, R.~Mazo, C.~Salinesi, and H.~B. Gh{\'e}zala,
  ``{Preference-based Conflict Resolution for Collaborative Configuration of
  Product Lines},'' in \emph{{International Conference on Evaluation of Novel
  Approaches to Software Engineering}}, Prague, Czech Republic, Mar. 2020.
  [Online]. Available: \url{https://hal.archives-ouvertes.fr/hal-02502398}
\BIBentrySTDinterwordspacing

\bibitem{Thum2014}
T.~Th\"{u}m, C.~K\"{a}stner, F.~Benduhn, J.~Meinicke, G.~Saake, and T.~Leich,
  ``{FeatureIDE: An Extensible Framework for Feature-Oriented Software
  Development},'' \emph{Science of Computer Programming}, vol.~79, p. 70–85,
  Jan. 2014.

\end{thebibliography}

\end{document}